\documentclass[letterpaper, 10pt, journal, twoside, final]{IEEEtran}  

\usepackage{units}
\usepackage{bm}
\usepackage{float}
\usepackage[font=small,skip=7pt]{caption}
\usepackage{epsfig}
\newlength\fwidth
\usepackage{mathtools}

\usepackage{xcolor}
\usepackage{siunitx}
\usepackage{multirow}
\usepackage{epstopdf}
\usepackage{amsmath} 
\usepackage{amssymb}  
\usepackage{mathtools}
\DeclareGraphicsExtensions{.pdf,.eps}
\usepackage[acronym,nomain,nonumberlist]{glossaries}
\newacronym{uav}{UAV}{Micro Aerial Vehicle}
\newacronym{nmpc}{NMPC}{Nonlinear Model Predictive Control}
\newacronym{mpc}{MPC}{Model Predictive Control}
\newacronym{panoc}{PANOC}{Proximal Averaged Newton-type method for Optimal Control}

\usepackage{url}

\begin{document}

\title{\LARGE \bf Nonlinear MPC for Collision Avoidance and Control of UAVs with Dynamic Obstacles}

  \author{Bj\"orn Lindqvist$^1$, Sina Sharif Mansouri$^1$, Ali-akbar Agha-mohammadi$^2$ and George Nikolakopoulos$^1$
  \thanks{Manuscript received: February 24, 2020; Revised June 1, 2020; Accepted July 1, 2020.}
  \thanks{This paper was recommended for publication by Editor N. Amato upon evaluation of the Associate Editor and Reviewers’ comments. This work has been partially funded by the European Unions Horizon 2020 Research and Innovation Programme under the Grant Agreement No. 730302 SIMS.}
  \thanks{$^{1}$The authors are with the Robotics and AI Team, Department of Computer, Electrical and Space Engineering, Lule\r{a} University of Technology, Lule\r{a} SE-97187, Sweden. First Author's email: \texttt{bjolin@ltu.se.}}%
  \thanks{$^{2}$The author is with Jet Propulsion Laboratory California Institute of
Technology Pasadena, CA, 91109. Third Author's email: \texttt{aliagha4@gmail.com.}}%
  \thanks{Digital Object Identifier (DOI): see top of this page.}}
\markboth{IEEE Robotics and Automation Letters. Preprint Version. Accepted July 1, 2020}%
{Lindqvist \MakeLowercase{\textit{et al.}}: Dynamic Collision Avoidance}

\captionsetup{font=footnotesize}
\maketitle
\begin{abstract}

This article proposes a Novel Nonlinear Model Predictive Control (NMPC) for navigation and obstacle avoidance of an Unmanned Aerial Vehicle (UAV). 
The proposed NMPC formulation allows for a fully parametric obstacle trajectory, while in this article we apply a classification scheme to differentiate between different kinds of trajectories to predict future obstacle positions.
The trajectory calculation is done from an initial condition, and fed to the NMPC as an additional input.  
The solver used is the nonlinear, non-convex solver Proximal Averaged Newton for Optimal Control (PANOC) and its associated software OpEn (Optimization Engine), in which we apply a penalty method to properly consider the obstacles and other constraints during navigation. The proposed NMPC scheme allows for real-time solutions using a sampling time of \unit[50]{ms} and a two second prediction of both the obstacle trajectory and the NMPC problem, which implies that the scheme can be considered as a local path-planner. This paper will present the NMPC cost function and constraint formulation, as well as the methodology of dealing with the dynamic obstacles. We include multiple laboratory experiments to demonstrate the efficacy of the proposed control architecture, and to show that the proposed method delivers fast and computationally stable solutions to the dynamic obstacle avoidance scenarios. 
\end{abstract}
\glsresetall 
\IEEEpeerreviewmaketitle

\section{Introduction}
\IEEEPARstart{N}{owadays}, Unmanned Aerial Vehicles (UAVs) demonstrate their promising capabilities in numerous application domains, such as infrastructure inspection~\cite{mansouri2018cooperative}, underground mine navigation~\cite{mansouri2019visioncnn}, search and rescue mission~\cite{tomic2012toward}, delivering first-aid or defibrillators in case of an accident~\cite{zegre2018delivery}, etc. In most of these scenarios, the interaction of the UAV with the environment is not considered and it is assumed that the surrounding obstacles remain static. However, to guarantee the overall safety and success of the mission and for demonstrating the capabilities of UAVs in close interaction with humans and in field trials, extending the classic notion of obstacle avoidance into the case of moving obstacles should be considered.

To be more specific, to deploy the UAVs in complex missions, such as urban environments, unstructured and unknown or constantly varying application scenarios with e.g. the existence of multiple moving obstacles and other robotic agents, it is essential to guarantee safe operations and zero incidents with inhabitants. In the urban environment, the UAV may interact with other aerial and ground vehicles, as well as  inhabitants, which are all constantly moving. Thus, this article proposes a novel collision avoidance approach for dynamic obstacles, included in the control layer, in order to guarantee a robust and online collision avoidance agile capability. As it will be presented, in the novel proposed framework, the predicted trajectory of the obstacle is fed to the Nonlinear Model Predictive Control (NMPC) for guaranteeing for a collision free path.
\subsection{Background and Motivation}
In the related scientific literature the majority of research has been done for the case of dynamic / moving obstacles, mainly in the context of an autonomous car ~\cite{fraichard1994car}, since it is required to ensure collision-free paths in the urban environment. These methods use a wide array of stochastic prediction models~\cite{althoff2009model} or hypothesis based models~\cite{ferguson2008detection} for the consideration of moving obstacles. 
For the case of an UAV, there have been few works focusing in the area of dynamic obstacle avoidance. The general consensus is to consider the obstacles / environment static and plan a path around it. For this approach, there is a myriad of different control approaches e.g. \cite{lavalle2006planning}. Reactive avoidance schemes include the widely used potential field methods~\cite{droeschel2016multilayered, kanellakis2018towards} and on-line graph-search methods, such as *ADA~\cite{heng2011autonomous}. 

The NMPC approach to path-planning has started to gain more traction in the field of UAVs as well~\cite{stastny2017nonlinear, erunsal2019decentralized}, once the issues of computation time was solved. These methods are often used in conjunction with other forms of obstacle avoidance, such as the potential fields~\cite{mansouri2019vision}, but there exist also examples of obstacle avoidance schemes using a pure NMPC structure~\cite{small2019aerial}. In \cite{kamel2017nonlinear} a NMPC was developed that considered a linear prediction model to ensure collision-free paths between agents in a collaborative scheme, but it is also paired with a potential field. The advantage of the NMPC scheme, over other obstacle avoidance schemes, is the ability to generate a collision-free trajectory based purely on the nonlinear kinematics of the UAV, where each point in the trajectory is described by a series of control inputs. In this specific application scenario, the main issue for the NMPC approach for UAVs, is the solver time. 

The adopted solver used in this article is the Proximal Averaged Newton for Optimal Control (PANOC)~\cite{sathya2018embedded, stella2017simple} and the associated open-source software OpEn (Optimization Engine) \cite{open2019}. As the name implies OpEn uses a Newton-type method and is specifically designed for optimal control problems, while it uses the same oracle as the projected gradient method \cite{nesterov2018lectures}, has a very low memory and computational footprint, and relies only on simple algebraic operations. OpEn also employs a penalty method~\cite{Hermans:IFAC:2018} for the consideration of equality constraints. The practicality and speed of OpEN makes it the prefect candidate for real-time applications of an NMPC scheme. 
%
\subsection{Contributions}
%
Based on the aforementioned state of the art, the first contribution of this article is in the coupling of the dynamic collision avoidance with the control layer, a concept that has never been presented before, to the authors best knowledge. In this novel approach, the NMPC is implemented for performing set-point tracking, while the nonlinear dynamic of the UAV is considered and proper constraint formulation allows the consideration of the dynamic obstacles. The overall proposed framework can be used as a baseline controller for guaranteeing collision avoidance and enables a larger application use for the UAVs. The second contribution is in defining the trajectory of the dynamic obstacles by classification and predicting the trajectory of an obstacle based on a discrete dynamics, feeding it directly to the optimizer as a parameter. In this way, the changes in the motion of the obstacle are considered in the prediction horizon of the NMPC formulation. The third contribution stems from the multiple laboratory experiments that display the efficacy of the proposed method for different scenarios of dynamic obstacle avoidance. This demonstration has significant novelty and impact as an enabler for a continuation of research efforts towards the real-life application of UAVs in dynamic environments.
\subsection{Outline} \label{sec:outline}
The rest of the article is structured as follows. Initially, the kinematic model of the UAV is presented in Section~\ref{sec:mavkinematic}, followed by the presentation of the corresponding objective function and the formulation of the obstacle constraints and trajectories in Sections~\ref{sec:obj} and~\ref{sec:const} respectively. A brief description of the optimization framework is presented in Section~\ref{sec:solver}, and the trajectory classification scheme is presented in \ref{sec:traj}. The experimental set-up and full control architecture is described in \ref{sec:setup}, while multiple experimental scenarios with related results and discussion that show the efficiency of the proposed framework are presented in Section~\ref{sec:experiments}. Finally, Section~\ref{sec:conclusion} concludes the article by summarizing the findings and offering directions for future research.
\section{Methodology} \label{sec:methodology}

\subsection{UAV Kinematics} \label{sec:mavkinematic}
\begin{figure}
\centering
\setlength{\belowcaptionskip}{-20pt}
  \includegraphics[width=0.7\linewidth]{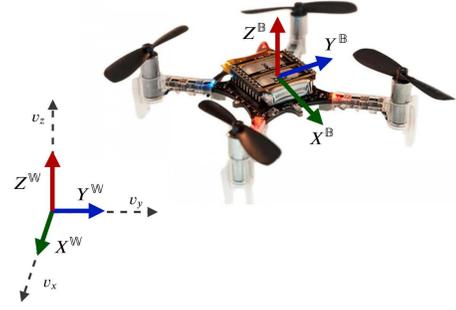}
  \caption{Utilized coordinate frames, where $\mathbb{W}$ and $\mathbb{B}$ denote the world and body coordinate frames respectively.}
  \label{fig:coordinateMAV}
\end{figure}

%
The UAV coordinate systems are depicted in Figure~\ref{fig:coordinateMAV}, where $(x^\mathbb{B}, y^\mathbb{B}, z^\mathbb{B})$ denote the body-fixed coordinate system, while $(x^\mathbb{W}, y^\mathbb{W}, z^\mathbb{W})$ denote the global coordinate system. In this article the states are defined in a yaw-compensated global frame of reference. The six degrees of freedom (DoF) for the UAV are defined by the set of equations \eqref{eq:mavkinematic}, while the full derivation of the adopted model can be found in~\cite{kamel2017model}. 

\begin{subequations}
\small
\label{eq:mavkinematic}
\begin{align}
        \dot{p}(t) &= v(t) \\ 
        \dot{v}(t) &= R(\phi,\theta) 
        \begin{bmatrix} 0 \\ 0 \\ T \end{bmatrix} + 
        \begin{bmatrix} 0 \\ 0 \\ -g \end{bmatrix} - 
        \begin{bmatrix} A_x & 0 & 0 \\ 0 &  A_y & 0 \\ 0 & 0 & A_z \end{bmatrix} v(t), \\ 
        \dot{\phi}(t) & = \nicefrac{1}{\tau_{\phi}} (K_\phi\phi_{\mathrm{ref}}(t)-\phi(t)), \\ 
        \dot{\theta}(t) & = \nicefrac{1}{\tau_{\theta}} (K_\theta\theta_{\mathrm{ref}}(t)-\theta(t)),
\end{align}
\end{subequations}
where $p=[p_x,p_y,p_z]^\top$ is the position and $v = [v_x,v_y,v_z]^\top$ is the linear velocity in the global frame of reference. $\phi$ and $\theta \in[-\pi,\pi]$ are the roll and pitch angles along the $x^\mathbb{W}$ and $y^\mathbb{W}$ axes respectively. Moreover, $R(\phi(t),\theta(t)) \in \mathrm{SO}(3)$ is a rotation matrix that describes the attitude in Euler form. $\phi_{\mathrm{ref}}\in \mathbb{R}$, $\theta_{\mathrm{ref}}\in \mathbb{R}$ and $T\geq 0$ are the reference inputs to the system in roll, pitch and the total thrust. The above model assumes that the acceleration depends only on the magnitude and angle of the thrust vector, produced by the motors, as well as the linear damping terms $A_x, A_y, A_z \in \mathbb{R}$ and the gravitational acceleration $g$. The attitude terms are modeled as a first-order system between the attitude (roll/pitch) and the references $\phi_{\mathrm{ref}}\in \mathbb{R}$, $\theta_{\mathrm{ref}}\in \mathbb{R}$, with gains $K_\phi, K_\theta\in\mathbb{R}$ and time constants $\tau_\phi, \tau_\theta \in \mathbb{R}$. These terms model the closed-loop behavior of a low-level controller, which also implies that the UAV is equipped with a lower-level attitude controller that takes thrust, roll and pitch commands and provides motor commands for the UAV.   
%

\subsection{Cost Function} \label{sec:obj}
%
We denote the state vector by $x = [p, v, \phi, \theta]^\top$, and the corresponding control actions as $u=[T,\phi_{\mathrm{ref}},\theta_{\mathrm{ref}}]^\top$.
The system dynamics of the UAV is discretized with a sampling time of $T_s$ using the forward Euler method to obtain $x_{k+1} = \zeta(x_k, u_k)$.
This discrete model is used as the \textit{prediction model} of the NMPC. The prediction is done with a receding horizon e.g., the prediction considers specified number of steps into the future. We denote this as the \textit{prediction horizon}, $N$, of the NMPC. By associating a cost to a configuration of states and inputs at the current time and in the prediction, a nonlinear optimizer is tasked with finding an optimal set of control actions, defined by the cost minimum of this \textit{cost function}. 
Let $x_{k+j{}\mid{}k}$ denote the predicted state 
at time step $k+j$, produced at the time step $k$. The corresponding
control actions are denoted by $u_{k+j{}\mid{}k}$.
Let us also denote $\bm{x}_{k}$ and
$\bm{u}_{k}$ the full predicted states and corresponding control inputs along the prediction horizon correspondingly. The goal of the controller is to make the states reach the prescribed set points and deliver smooth control inputs. Thus we formulate the cost function as:
\begin{multline}
\label{eq:costfunction}
J(\bm{x}_{k}, \bm{u}_{k}, u_{k-1\mid k}) = \sum_{j=0}^{N}   \underbrace{\| x_{\mathrm{ref}}-x_{k+j{}\mid{}k}\|_{Q_x}^2}_\text{State cost} 
\\
+   \underbrace{\| u_{\mathrm{ref}}-u_{k+j{}\mid{}k}\|^2_{Q_u}}_\text{Input cost}
+  \underbrace{\| u_{k+j{}\mid{}k}-u_{k+j-1{}\mid{}k} \|^2 _{Q_{\Delta u}}}_\text{Input smoothness cost},
\end{multline}
where $Q_x\in \mathbb{R}^{8\times8}, Q_u, Q_{\Delta u}\in 
\mathbb{R}^{3\times3}$ are positive definite weight matrices for the
states, inputs and input rates respectively. In \eqref{eq:costfunction}, the 
first term denotes the \textit{state cost}, which penalizes deviating from a 
certain state reference $x_{\mathrm{ref}}$. The second term denotes the 
\textit{input cost} that penalizes a deviation from the steady-state input 
$u_{\mathrm{ref}} = [g, 0, 0]$ i.e. the inputs that describe hovering. Finally,
to enforce smooth control actions, a third term is added that penalizes changes
in successive inputs. Note that the first such penalty, $\| 
u_{k{}\mid{}k}-u_{k-1{}\mid{}k} \|^2$, depends on the previous
control action.
%

\subsection{Obstacle Definition and Constraints} \label{sec:const}
%
Following the constraint formulation structure for OpEn used in \cite{small2019aerial}, while also keeping the constraints fully parametric so that their positions and size is part of the input fed to the NMPC scheme, we use the function  $[h]_+=\max\{0, h\}$. This allows us to formulate the constraints as equality expressions such that $[h]_+=0$ implies that the constraint is satisfied. We can use this formulation to express a constrained area by choosing \textit{h} as an expression that is positive inside of the constrained area and negative outside of it. For the experimental scenarios described in \ref{sec:experiments} a simple spherical obstacle fits very well to the needs. 
 Moreover, to guarantee bounds on changes in control actions, a constraint on the control input variations will be also established. Finally, it should be highlighted that all the underlying constraints are considered in the full control horizon \textit{N} to account for the constraints at all the predicted future time steps.

\subsubsection{Spherical Obstacle} 
The spherical obstacle represents the dynamic obstacle that is thrown at/approaching the UAV, with an included safety distance. For now the obstacle is defined with an arbitrary trajectory, i.e. the position of the sphere at each time step in the prediction is set as an input to the solver. Taking inspiration from the collision avoidance scheme in \cite{kamel2017nonlinear} we increase the radius of the obstacle along the prediction horizon. This is needed due to the fact that with imperfect measurements and prediction models, the further the prediction is, the larger the errors in the obstacle trajectory will be, which is unavoidable. The obstacle constraint is defined as:
\begin{multline}\label{eq:sphericalconstraint} 
    h_{\mathrm{sphere}}(p, \xi^{\mathrm{obs}}) = [(r^{\mathrm{obs}} + r_{\mathrm{s}})^2 - (p_x{-} p_x^{\mathrm{obs}})^2 \\
    - (p_y{-}p_y^{\mathrm{obs}})^2 - (p_z{-}p_z^{\mathrm{obs}})^2]_+ = 0,
\end{multline}
where $\xi^{\mathrm{obs}} = [r_{\mathrm{obs}}, r_{\mathrm{s}}, p_x^{\mathrm{obs}}, p_y^{\mathrm{obs}}, p_z^{\mathrm{obs}}]$. The obstacle positions $p^\mathrm{obs}$ are the world-frame coordinates of the center of the sphere, $r_{\mathrm{obs}}$ is the radius of the obstacle and $r_{\mathrm{s}}$ is an extra safety radius. This implies that as long as the UAV position $p$ lies outside the sphere \eqref{eq:sphericalconstraint} is zero and can thus be stated as an equality constraint. To extend this notion to a dynamic obstacle, we form a separate constraint for each predicted time step. Let $\bm{p}_x^{\mathrm{obs}}, \bm{p}_y^{\mathrm{obs}}, \bm{p}_z^{\mathrm{obs}}$ denote the vectors of size $N$ that describe the obstacle trajectory, $\bm{r}_s$ denote a linearly increasing safety radius along $N$, and $\bm{p}_k$ denote the full predicted UAV positions at time step $k$. Then the vector of obstacle constraints can be formulated as:
\begin{equation}\label{eq:vector_const}
    \bm{h}_{\mathrm{sphere}}(\bm{p}_k, \bm{\xi}^{\mathrm{obs}}) = \overline{0},
\end{equation}
where $\bm{\xi}^{\mathrm{obs}}= [r^{\mathrm{obs}}, \bm{r}_{\mathrm{s}}, \bm{p}_x^{\mathrm{obs}}, \bm{p}_y^{\mathrm{obs}}, \bm{p}_z^{\mathrm{obs}}]$, such that the constraints are satisfied if the UAV does not enter the spherical obstacle, as described by the obstacle trajectory, at any of the predicted UAV positions. As such, the trajectory of the obstacle is fully described and parametrized by $\bm{\xi}^{\mathrm{obs}}$. The NMPC is not set for a specific trajectory of the obstacle, and for the identification and prediction of the dynamic obstacle's trajectories the constraint-formulation is agnostic to the method of trajectory prediction.

\subsubsection{Control Input Rate}
Rapidly moving out of the trajectory of an incoming obstacle it can result in an aggressive or oscillatory behavior of the control inputs and thus we impose a constraint on the successive differences of control actions. The purpose of this constraint is to set a bound on the magnitude of the change in control inputs $\phi_{\mathrm{ref}}$ and $\theta_{\mathrm{ref}}$, which is done by an upper and a lower bound. This can be written as an equality constraints as:
\begin{subequations}\label{eq:delta_constraints}
\begin{align}
    [\phi_{\mathrm{ref}, k+j-1{}\mid{}k} 
    - \phi_{\mathrm{ref},k+j{}\mid{}k}
    -\Delta \phi_{\max}]_+ {}={}& 0,
    \\
    [\phi_{\mathrm{ref},k+j{}\mid{}k}
    -\phi_{\mathrm{ref}, k+j-1{}\mid{}k}
    -\Delta \phi_{\max}]_+ {}={}& 0.
\end{align}
\end{subequations}
Additionally we also form the same constraint for $\theta$, with $\Delta \phi_{\max}$ and $\Delta \theta_{max}$ as the maximum change in input per time step.
\subsubsection{Input constraints} \label{sec:input_constraints}
Finally, we also directly apply constraints on the control inputs. Since the NMPC is to be used with a real UAV, hard bounds on reference angles $\phi_{\mathrm{ref}}, \theta_{\mathrm{ref}}$ must be considered, as a low-level controller will only be able to stabilize the attitude within a certain range. Since the thrust of a UAV is limited, such hard bounds must also be applied to the thrust input, \textit{T}. Thus we define bounds on inputs as:
\begin{equation}
\label{eq:input_const}
u_{\min} \leq u_{k+j\mid k} \leq u_{\max}.
\end{equation}
\subsection{Optimization} \label{sec:solver}
The NMPC problem is solved by PANOC \cite{sathya2018embedded,small2019aerial,open2019}, while a penalty method is applied for the consideration of equality constraints. OpEn solves general parametric optimization problems on the form:
 \begin{subequations}\label{eq:open_problem}
\begin{align}
    \operatorname{Minimize}_{z \in Z}\,& f(z,\rho)
    \\
    \text{subject to:}\,& F(z,\rho) = 0,
\end{align}
\end{subequations}
where $f$ is a Lipschitz-differentiable function and $F$ is a vector-valued mapping so that $\|F(z,\rho)\|^2$ is a Lipschitz-differentiable function. The decision variable and parameter are denoted by $z$ and $\rho$ respectively. 
Based on the cost function and constraints outlined in \ref{sec:obj} and \ref{sec:const} we can formulate the NMPC problem with $N_s$ spherical obstacles with parameterized trajectories, as:
 \begin{subequations}\label{eq:nmpc}
\begin{align}
    \operatorname*{Minimize}_{
        \bm{u}_k, \bm{x}_k
    } \,
    & J(\bm{x}_{k}, \bm{u}_{k}, u_{k-1\mid k})
    \\
    \text{s. t.:}\,& 
    x_{k+j+1\mid k} = \zeta(x_{k+j\mid k}, u_{k+j\mid k}),\notag
     \\ & j=0,\ldots, N-1,
    \\
    &u_{\min} \leq u_{k+j\mid k} \leq u_{\max},\, j=0,\ldots, N,
\label{eq:nmpc:input_constraints}
    \\
    &h^i_{\mathrm{sphere}}(p_{k+j\mid k}, \xi^{\mathrm{obs},i}_j) = 0,\,
     j=0,\ldots, N, \\
     &i = 1,\ldots, N_s
     \\
     &\text{Constraints \eqref{eq:delta_constraints}},
     j=0,\,\ldots, N.
\end{align}
\end{subequations}

This can be fit into the OpEn framework by performing a \textit{single-shooting} of the cost function via decision variable $z = \bm{u}_k$ and define $Z$ by the input constraints \eqref{eq:input_const}. We also define $F$ to cast the equality constraints \eqref{eq:sphericalconstraint} and \eqref{eq:delta_constraints}. The parameter $\rho$ is set to include initial conditions, references and the obstacle trajectory. For the consideration of the equality constraints, a quadratic penalty method is applied. By formulating the problem as
\(
    \operatorname{Minimize}_{z \in Z} f(z,\rho) + c\|F(z,\rho)\|^2
\),
where $c$ is a positive penalty parameter, the PANOC algorithm can be applied to the problem. In the penalty method an optimization problem, where the constraints are mapped to the cost-domain, is re-solved multiple times with an increasing penalty parameter $c$ associated to the constraints, while using the previous solution as the initial guess. This method gradually moves the cost-minima until non of the constraints are violated, or rather until a specified tolerance is met. In \ref{sec:experiments} a maximum of four penalty method iterations are applied. 
\subsection{Trajectory Classification}\label{sec:traj}
While many different types of trajectories of obstacles can be encountered in the urban environment, and the NMPC formulation allows for trajectories of arbitrary shape, in this article we will limit the trajectories to three categories, namely: linear motion, projectile motion, or static obstacles. As such, a direct prediction model to predict the future positions of the obstacle could be utilized in case that it is possible to compute its \textit{trajectory class}. General movements, such as pedestrians or cars in the urban environment, are often moving from starting point to destination, i.e. linear movement, while objects thrown or rock-falls follow the projectile motion equations. Additionally we include a static obstacle class, without any movement ($\dot{p}^{obs} = 0$). 
Linear motion is described by:
\begin{equation}\label{eq:linear}
   \dot{p}^{\mathrm{obs}}(t) = v^{obs}(t),
\end{equation}
where $v^{obs}$ are the velocities of the obstacle and no forces are acting on the obstacle. The projectile-motion trajectory is defined by:
\begin{subequations}
\label{eq:projectile}
\begin{align}
   \dot{p}^{\mathrm{obs}}(t) &= v^{obs}(t),
   \\
   \dot{v}^{\mathrm{obs}}(t) &= \begin{bmatrix} 0 \\ 0 \\ -g \end{bmatrix} - 
   \begin{bmatrix} B_x & 0 & 0 \\ 0 &  B_y & 0 \\ 0 & 0 & B_z \end{bmatrix} v^{obs}(t),
\end{align}
\end{subequations}
where, much like in \eqref{eq:mavkinematic}, $B$ are linear aerodynamic damping terms. The buoyancy force of the obstacle is considered small enough to ignore. Equations \eqref{eq:projectile},~\eqref{eq:linear} are then discretized by forward Euler using the same sampling time as the controller/prediction model, $T_s$. For the obstacle states $x^{\mathrm{obs}} = [p^{\mathrm{obs}}, v^{\mathrm{obs}}]$, we thus have the discrete-predictive form 
\begin{equation}\label{eq:obspred}
    x^{obs}_{k \mid k + n + 1} = \alpha_t(x^{obs}_{k \mid k + n}),
\end{equation}
where $\alpha_t$ denoted the discrete prediction model of the trajectory, and can be iterated indefinitely. Similarly for the backwards prediction of $x^{obs}_{k \mid k + n - 1} = \beta_t(x^{obs}_{k \mid k + n})$. 
In the discrete prediction model of the projectile-motion we also include a condition for bouncing, with a much-simplified collision interaction with a coefficient of restitution applied on the velocities (assuming the ground is flat) as the sphere hits the ground to model the energy-loss in the collision.

The classification is done by comparing the \textit{M} last measured position and velocity terms to a backwards prediction based on \eqref{eq:linear} and \eqref{eq:projectile} iterating from the current measured state, $x^{obs}_{k \mid k}$. Using the same notation as for the prediction of the NMPC, as $x^{obs}_{k \mid{} k }$ denotes the current measurement let $x^{obs}_{k \mid{} k - j}$ denote the predicted obstacle state $j$ steps back in time.
Also denote the vector of previous measurements in position and velocity of the obstacle as $\bm{p}^{prev}_j$ and $\bm{v}^{prev}_j$ respectively. 
We measure the error, $e^{traj}$, as:
\begin{equation}\label{eq:error}
  e^{traj} = \sum_{j=1}^{M} \mid{} p^{prev}_j - p^{obs}_{k \mid{} k - j} \mid{} + \mid{} v^{prev}_j - v^{obs}_{k \mid{} k - j} \mid{}. 
\end{equation}
Equation \eqref{eq:error} is evaluated for the three different classes of trajectories and the class that generates the lowest error is chosen for the trajectory prediction. This is run for every new measurement of the obstacle and thus the trajectory of a single obstacle is allowed to change during the movement, and the predicted trajectory only depends on the current measurement. 
We can then form the full obstacle parameter $\bm{\xi}^{\mathrm{obs}}= [r^{\mathrm{obs}}, \bm{r}_{\mathrm{s}}, \bm{p}^{\mathrm{obs}}]$ by computing $\bm{p}^{\mathrm{obs}} \in \mathbb{R}^{3 \times N}$ from $x^{obs}_{k \mid{} k }$ by iterating the discrete prediction model in \eqref{eq:obspred} to produce $[p^{\mathrm{obs}}_{k \mid k}, p^{\mathrm{obs}}_{k \mid k+1}, \ldots p^{\mathrm{obs}}_{k \mid k + N}]$. The radii $r^{\mathrm{obs}}, \bm{r}_{\mathrm{s}}$ are, in this paper, provided by the operator. 
%
\section{Results} \label{sec:results}
\subsection{Experimental Set-up} \label{sec:setup}
\begin{figure*}[ht]
\centering
\setlength{\belowcaptionskip}{-15pt}
  \includegraphics[scale=0.36]{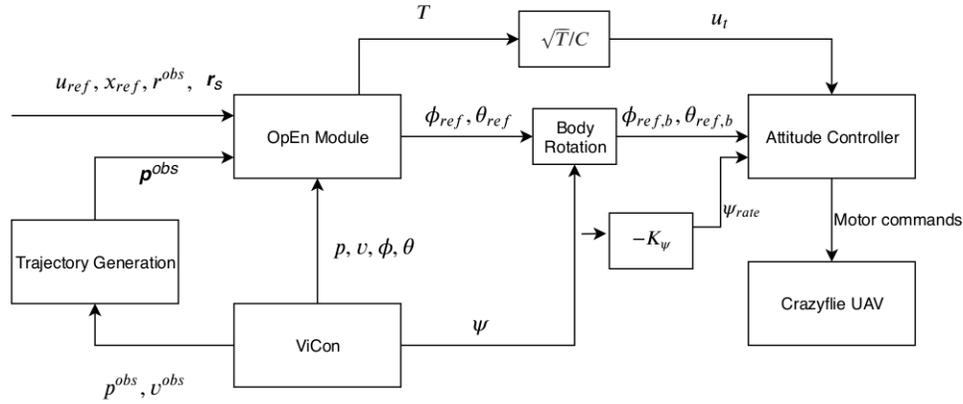}
  \caption{Proposed Control Architecture. The Vicon motion capture system (and a median filter for velocity estimation from the position data) provides state data for the UAV $(p,v,\theta,\phi)$  and the obstacle $(p^{\mathrm{obs}},v^{\mathrm{obs}})$. The obstacle data is used to classify the trajectory and generate a predicted trajectory based on the measured initial condition. The UAV state data and the obstacle trajectory is fed to the NMPC module as the solver parameter which also include $x_{\mathrm{ref}}, u_{\mathrm{ref}}, \bm{r}_s, r^{\mathrm{obs}}$. The NMPC generates control inputs $\phi_{\mathrm{ref}}, \theta_{\mathrm{ref}}$ and $T$, which after the relevant mapping are fed to the low-level attitude controller.}
  \label{fig:architeture}
\end{figure*}

For all the following experiments we use a Vicon Motion-Capture System to track the UAV and the obstacle. All the state data used in the NMPC, namely the UAV $(p,v,\theta,\phi)$ and the obstacle $(p^{\mathrm{obs}},v^{\mathrm{obs}})$, are provided by Vicon using a complementary filter to estimate the velocities. The chosen platform is the Crazyflie 2.0 Nano Quadcopter (and is also seen in Figure \ref{fig:coordinateMAV}), which is a small and lightweight quadcopter. This type of smaller platform was chosen for safety reasons due to the nature of the experiments where obstacles are thrown at the UAV. The Crazyflie has no on-board computer and thus all the computation is done on a remote laptop, namely a Lenovo 430 Thinkpad with a 3rd gen Core i5 \unit[2.60]{GHz} and 4GB of RAM. To allow communication and low-level attitude control of the Crazyflie we use the open source Robot Operating System (ROS)~\cite{quigley2009ros} package developed for the Crazyflie~\cite{crazyflieROS}. 
As can be seen in Figure \ref{fig:architeture} the trajectory is generated outside of the NMPC module and fed to the optimizer as an additional input, together with other user-specified parameters. We also perform a basic thrust-mapping, since \cite{crazyflieROS} accepts an input $u_t \in [0,1]$, where we assume $u_t$ has a quadratic relation to the thrust. The low-level controller also accepts a yaw-rate command, and as seen in Figure \ref{fig:architeture} we use a basic P-controller with gain $K_{\psi}$ to keep the yaw, $\psi$, at zero.
Despite forcing the yaw to zero we also covert the control inputs from global to local frame. 

\subsection{Laboratory Experiments} \label{sec:experiments}
For the presented experiments, the corresponding model parameters that are described as in \eqref{eq:mavkinematic} are chosen as $\tau_{\phi}, \tau_{\theta} = 0.5$, $K_{\phi}, K_{\theta} = 1$, in order to approximately match the response of a low-level controller acting on an UAV, while damping terms are set to $A_x = 0.1, A_y = 0.1, A_z = 0.2$. Additionally, \textit{g} is set to $\unitfrac[9.81]{m}{s^2}$, the control horizon is set to $N = 40$ and $T_s = \unit[50]{ms}$ (the same as in \cite{small2019aerial}),  which implies a prediction of two seconds. The selection of horizon $N$ is based on the trade-off between computational complexity and considering the obstacle in time to avoid collision. The prediction horizon (in combination with $T_s$) is the limiting factor to how early the system can react to incoming obstacles.  The weights in \eqref{eq:costfunction} are chosen as $Q_x = \operatorname{diag}(5,5,30,3,3,3,8,8)$, $Q_u = \operatorname{diag}(5, 10, 10)$, $Q_{\Delta u} = \operatorname{diag}(5, 12, 12)$, while the  constraints on control inputs are (in SI-units) as $u_{min} = [5.0, -0.35, -0.35]^\top$ and $u_{max} = [13.5, 0.35, 0.35]^\top$.

Additionally the constraints on change in input described in \eqref{eq:delta_constraints} are chosen as $\Delta\phi_{max} = 0.08$ and $\Delta\theta_{max} = 0.08$. The safety radius $\bm{r}_s$ is set to \unit[0.0]{m} at $j=0$ and is increased linearly to \unit[0.2]{m} at $j=N$, while the obstacle radius $r^{obs}$ is specified for each experiment. The classification is done with the number of backwards measurements $M$ = 5. 

Due to the nature of the following experiments, with multiple simultaneously moving objects, we recommend also watching the video summary of the experiments to get a more clear view of the set-up and results. The video includes a brief comparison with other obstacle avoidance methods, and the four experimental scenarios discussed in the sequel. It can be found at: \textbf{\url{https://youtu.be/vO3xjvMMNJ4}}. Each experiment using the proposed method was performed 5-10 times without collisions, and the data presented in this section and in the video is based purely on the best-looking trajectories.

\subsubsection{Position hold while avoiding projectile}
The task of the UAV is to hold position while avoiding any incoming obstacles, which is a projectile thrown at the UAV. We will use this scenario to evaluate how popular methods, that do not include a trajectory prediction, performs at avoiding an incoming projectile. While many methods such as the artificial potential fields have been evaluated for dynamic obstacles \cite{budiyanto2015uav}, it is under the assumption that the UAV is fast enough to respond to the moving obstacle while still considering it static (but updating its position on-line). Since we are only interested in avoiding one point-like incoming projectile we form the repulsive force as $F_{rep} = L (1 - \frac{\mid \mid p' \mid \mid}{d_s})^2\frac{p'}{\mid \mid p' \mid \mid}$, where $p' = (p - p^{\mathrm{obs}})$, $d_s$ is the radius of influence of the potential field, and $L$ is the repulsive constant which is chosen very large for an aggressive response, while the attractive force is the hold-position position reference. We also compare with the same NMPC-constraint method presented in this paper (being similar to \cite{small2019aerial}, using a spherical obstacle), but considering the obstacle static. To assist these methods we choose the radius of influence and obstacle radius respectively to $1m$, which is much larger than the needed safety distance to avoid a static obstacle, while tuning the reference-following controller of the potential field to be as aggressive as possible, also relaxing input-rate constraints \eqref{eq:delta_constraints} for the NMPC for a faster response. 

While maintaining distance to a slowly moving obstacle (also shown in the video), both these methods fail at avoiding collisions with the projectile-motion obstacle as shown in Figure \ref{fig:dist_all}. There is simply not enough time from the obstacle entering the area of influence of these controllers until collision with the UAV, to initiate avoidance in time, nor do the controllers have a notion of where the obstacle's future position are and as such their avoidance maneuver might move them along the trajectory of the obstacle. 

\begin{figure}[ht]
\centering
\setlength{\belowcaptionskip}{-10pt}
  \includegraphics[width=\linewidth]{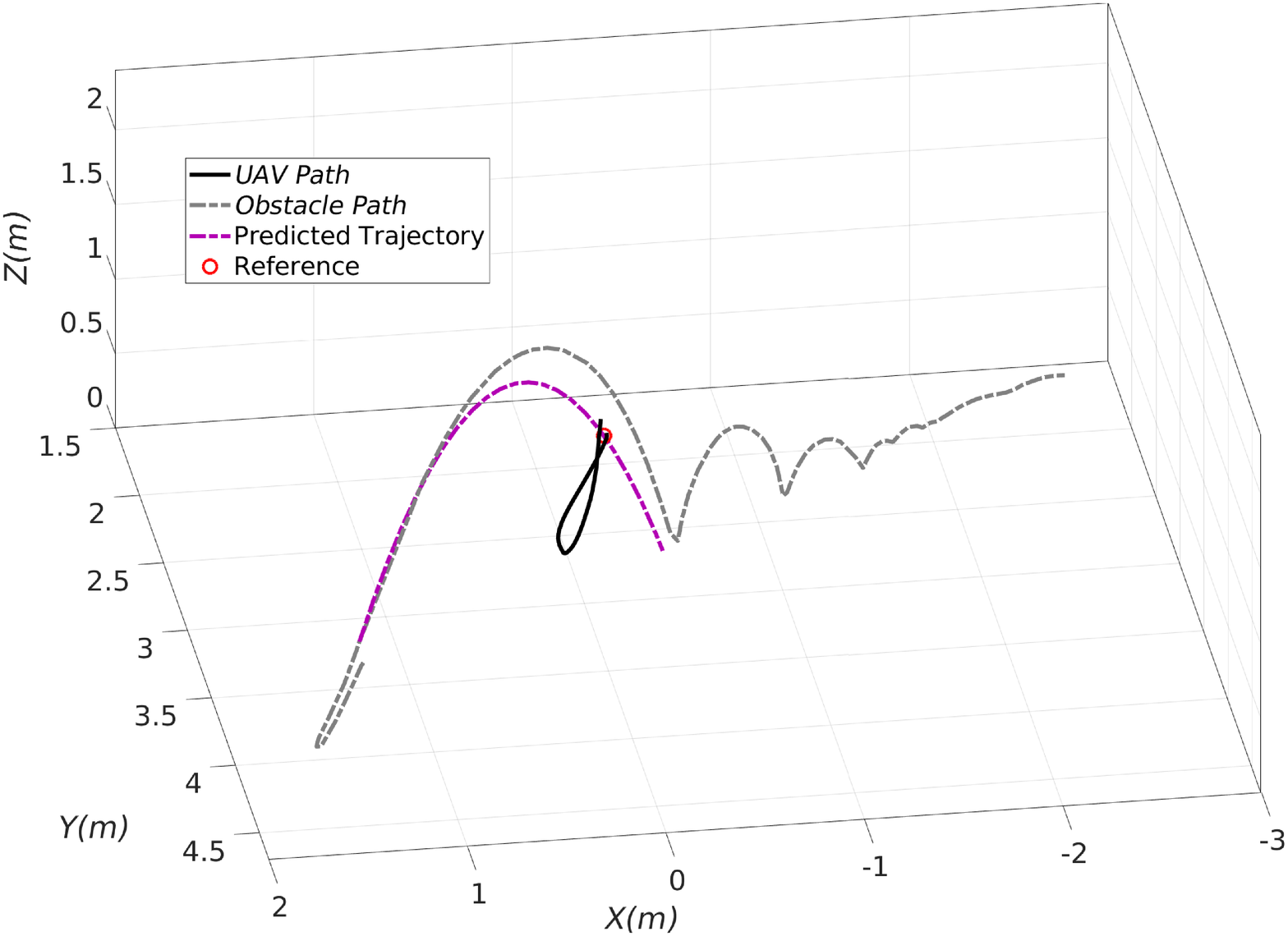}
  \caption{Path of UAV and dynamic obstacle during projectile-motion experiment.}
  \label{fig:proj_path}
\end{figure}

\begin{figure}[ht]
\centering
\setlength{\belowcaptionskip}{-10pt}
  \includegraphics[width=\linewidth]{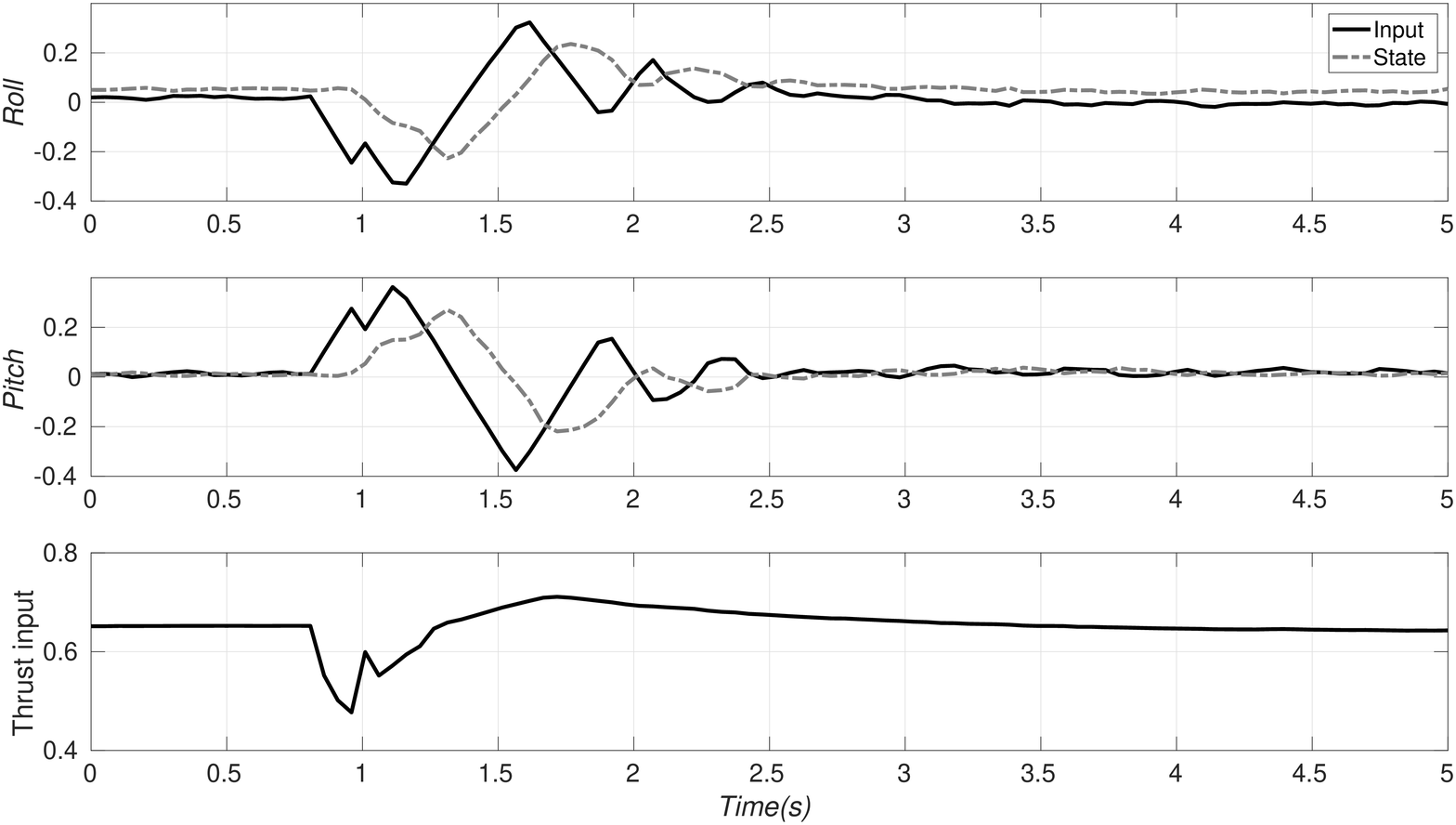}
  \caption{Control inputs during projectile-motion experiment.}
  \label{fig:proj_inputs}
\end{figure}

Using the predictive method and trajectory classification described in this paper and setting the obstacle radius to \unit[0.4]{m}, the approximate distance required to safely not result in a collision, we test the proposed controller in the same scenario. Figure \ref{fig:proj_path} shows the paths of the projectile and UAV. The obstacle is thrown at approximately \unit[0.7]{s}, as seen by the relative distances in Figure \ref{fig:dist_all}, and the avoidance maneuver starts at approximately \unit[0.8]{s} in Figure \ref{fig:proj_inputs}. Figure \ref{fig:proj_path} also displays the predicted trajectory at the moment the avoidance maneuver starts (note: the bounce condition is not applied here, see \ref{sec:bounce}). 

\begin{figure}[ht]
\centering
\setlength{\belowcaptionskip}{-5pt}
\includegraphics[width=\linewidth]{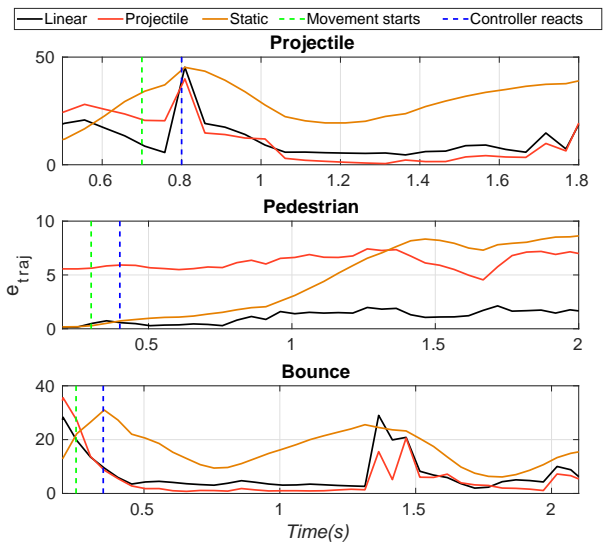}
  \caption{The prediction error, $e_{traj}$ for the dynamic obstacle, during relevant time samples for the first three three experiments. The class with the lowest error is chosen for the prediction of the obstacle trajectory.}
  \label{fig:etraj}
\end{figure}

The prediction error of the obstacle, used for trajectory classification, can be found in Figure \ref{fig:etraj}, as well as the point where the predicted trajectory is determined to be on a collision course with the UAV (within the prediction horizon), which is also when the controller initiates the avoidance. We can see that the NMPC computes control inputs to avoid the obstacle as soon as the trajectory is classed as projectile-motion. The classification successfully identifies the projectile-motion trajectory and the obstacle avoidance is initiated in time to successfully avoid the incoming projectile. Figure \ref{fig:solvertime_all} displays the solver time, which peaks at \unit[40]{ms}. The minimum distance in this experiment was \unit[0.51]{m} (due to $\bm{r}_s$). Although not a perfect hit, the minimum distance of the initial position of the UAV and the projectile was around \unit[0.17]{m} which would result in a collision since the distance is calculated from the center of the obstacle to the center of the UAV.

At exactly one second into the experiment the trajectory classification fails for one instant, as seen in Figure \ref{fig:etraj}, and classifies the trajectory as linear, and can be seen in Figure \ref{fig:dist_all} where the penalty method iterations are not applied and such the solver time drastically decreases, since the obstacle prediction is no longer on a collision course.
In Figure \ref{fig:proj_inputs} a problem regarding remotely controlling the UAV is also shown, and will persist in the upcoming experiments as well. There is a delay of approximately \unit[0.1 - 0.15]{s} between the control input and the state changing, which is the nature of a remotely controlled UAV and can not be avoided.  

\subsubsection{Position hold while avoiding pedestrian}
\begin{figure}[ht]
\centering
\setlength{\belowcaptionskip}{-10pt}
  \includegraphics[width = \linewidth]{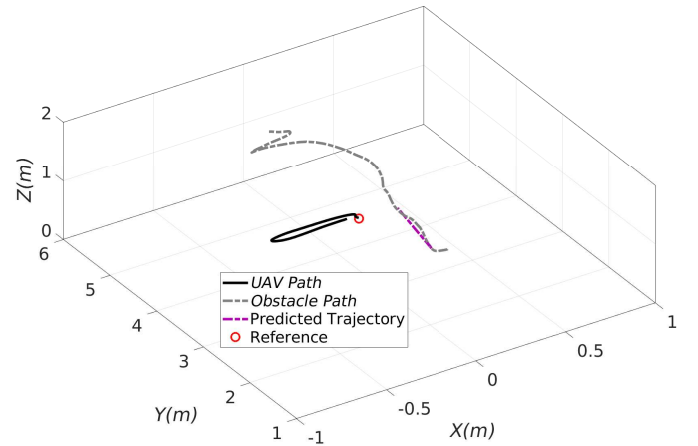}
  \caption{Path of UAV and dynamic obstacle during linear motion.}
  \label{fig:linear_path}
\end{figure}

The task is to hold a position, while avoiding approaching obstacles. In this experiment a "pedestrian" is walking towards the UAV on a direct collision course. The obstacle radius is set to \unit[0.6]{m}, due to the larger size of the incoming obstacle. The path of the UAV and obstacle is shown in Figure \ref{fig:linear_path}, as well as the predicted trajectory starting from the time step that the trajectory is identified and within the prediction horizon, and the avoidance maneuver starts. As in the previous case, looking at Figure \ref{fig:dist_all} and Figure \ref{fig:etraj} the pedestrian starts to move at \unit[0.3]{s} into the experiment, while the controller reacts to initiate the avoidance maneuver  at around \unit[0.4]{s}. Figure \ref{fig:dist_all} shows that the minimum distance is \unit[0.59]{m}, which implies that the UAV cleanly avoids the collision. The solver time is also found in Figure \ref{fig:solvertime_all} and peak to a maximum of \unit[29]{ms}. 

\subsubsection{Bounce condition}\label{sec:bounce}
\begin{figure}[ht]
\setlength{\belowcaptionskip}{-5pt}
  \includegraphics[width = \linewidth]{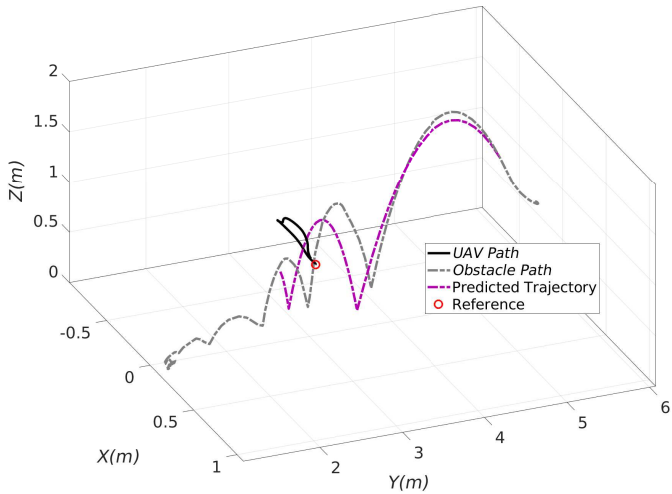}
  \caption{Path of UAV and dynamic obstacle during experiment with bouncing ball.}
  \label{fig:bounce_path}
\end{figure}

The bouncing ball is often seen as an interesting hybrid system. Thus to show the power of this type of NMPC structure, where the trajectory of the obstacle is provided by an external source, we include an experiment of the avoidance of a bouncing ball. As in the previous two cases the task of the UAV is to hold position, while avoiding incoming obstacles, and the obstacle radius is set to \unit[0.4]{m}. In this experiment the obstacle is thrown on a trajectory such that it would impact the UAV after the first bounce (approximately \unit[2.1]{s} the experiment in Figure \ref{fig:dist_all}). The obstacle is thrown at approximately \unit[0.25]{s}, while the controller reacts at \unit[0.35]{s}, also seen in Figure \ref{fig:etraj}. This displays that even a simplified trajectory model still results in a good enough prediction of the obstacle path, especially with the inclusion of an increasing safety radius along the prediction. The predicted trajectory of the obstacle, based on the initial condition at the time when the avoidance maneuver starts, is displayed in Figure \ref{fig:bounce_path}, together with the measured path of the obstacle and UAV. The UAV successfully avoids the obstacle with a minimum distance of \unit[0.38]{m} which can be seen in Figure \ref{fig:dist_all}, while the solver time peaks to \unit[33]{ms}. Such a small constraint violation is expected due to solver tolerances and imperfect measurements.

\subsubsection{Multiple Obstacles}
\begin{figure}[ht]
\centering
\setlength{\belowcaptionskip}{-5pt}
  \includegraphics[width = \linewidth]{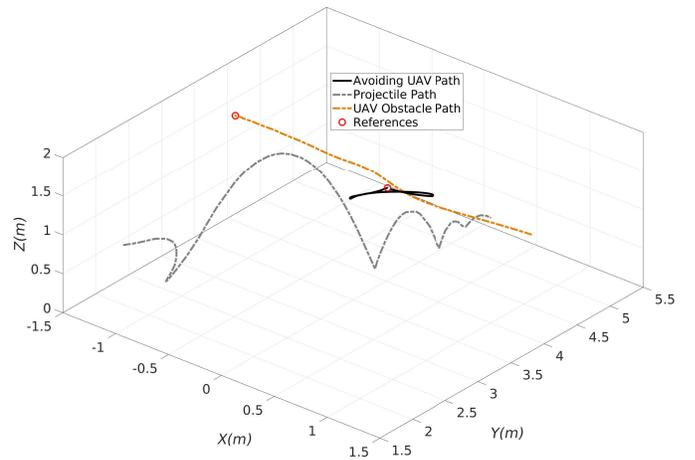}
  \caption{Paths of (avoiding) UAV, projectile-motion obstacle, and UAV obstacle.}
  \label{fig:multiple_path}
\end{figure}

\begin{figure}[ht]
\centering
\setlength{\belowcaptionskip}{-5pt}
  \includegraphics[width=\linewidth]{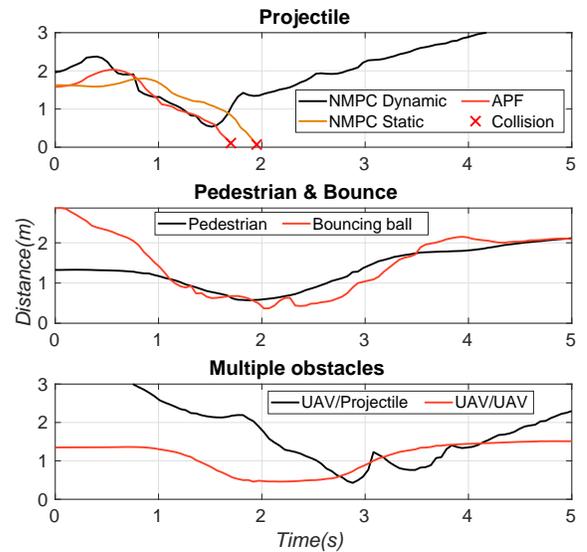}
  \caption{Distance measurements between the UAV and obstacles for the comparisons and the four described obstacle avoidance scenarios.}
  \label{fig:dist_all}
\end{figure}

\begin{figure}[ht]
\centering
\setlength{\belowcaptionskip}{-5pt}
  \includegraphics[width=\linewidth]{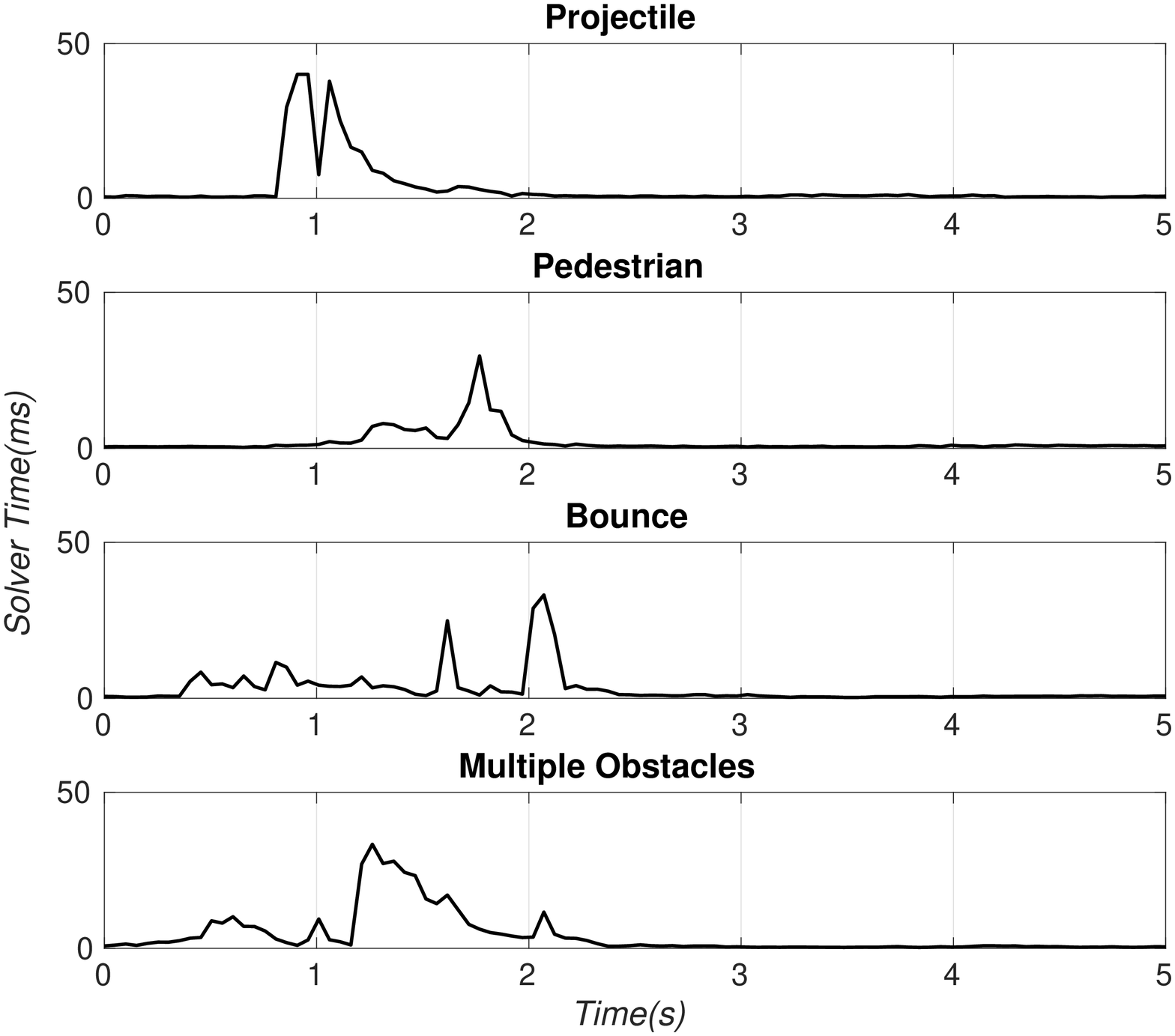}
  \caption{Solver time of the NMPC for the four obstacle avoidance scenarios.}
  \label{fig:solvertime_all}
\end{figure}

Finally, we evaluate the proposed method in terms of avoiding multiple dynamic obstacles. The experimental set-up is such that a separate UAV is set on a collision course with the avoiding UAV, while a projectile is simultaneously thrown at it, both set with $r^{\mathrm{obs}} = 0.4$. The trajectory classification and prediction scheme is applied on separate measurements of the two obstacles, but is otherwise used identically as in the single obstacle cases. The trajectories of the two UAVs and the projectile are shown in Figure \ref{fig:multiple_path}, while the distances between the avoiding UAV and the approaching UAV and obstacle are shown in Figure \ref{fig:dist_all}. The safety distances are maintained, with a minimum distance of \unit[0.45]{m} and \unit[0.42]{m} respectively for the two incoming obstacles. It should be noticed also how the avoiding UAV keeps the safety distance for a prolonged time to the approaching UAV while it is also avoiding the incoming projectile. In the video results, the avoidance maneuver can also be seen to start as soon as the obstacle-UAV starts its motion. The solver time in Figure \ref{fig:solvertime_all} peaks at \unit[35]{ms}, which is similar to the single-obstacle case.

%
\section{Conclusions} \label{sec:conclusion}

In this paper we proposed a novel path-planning and obstacle avoidance scheme that has the ability to handle dynamic obstacles using a NMPC architecture. To supplement this we also include a simple trajectory classification scheme to better display the applicability of the controller in a more general approach, so that the same scheme can be used in all experimental scenarios. The proposed scheme of NMPC and trajectory classification successfully provides collision-free paths in all the considered cases. The online optimization problem is solved within the required real-time restrictions of \unit[50]{ms} without violation of the established obstacle or input constraints. The limitations of this method are: a) the fact that the overall performance is based on the reliance on classification of trajectories, and b) using an explicit prediction of future obstacle positions. If the predictive scheme fails or involve too large errors, the UAV might completely ignore an obstacle on a collision-course and even for our limited study of trajectories our scheme momentarily classifies the trajectory incorrectly. 

This paper displays a novel and interesting approach to obstacle avoidance of UAVs, that can be much expanded on by further research. This future work includes more general trajectory identification such as curve-fitting or estimating the forces acting on the obstacle, and of course removing the reliance of a motion-capture system for obstacle detection. The last part would include extracting the position and velocity of obstacles using, for example, stereo-cameras or 3D lidars. Although the trajectory classification scheme never failed in a way that resulted in collisions in the around 40 performed experiments, further statistical analysis on how such a scheme performs in more difficult scenarios should also be considered.

Additionally, further analysis on how the complexity of the NMPC problem scales with more obstacles and how that relates to the solver time, to see at what point it is more appropriate to not solve for obstacle avoidance directly in the control layer. This type of control structure is, in the authors opinion, an interesting path towards using UAVs in urban dynamic environments or any environment where collision avoidance is of great importance to ensure the safety of persons and vehicles. 
\bibliography{mybib}
\end{document}